\documentclass{article}

\usepackage{microtype}
\usepackage{graphicx}
\usepackage{subfigure}
\usepackage{booktabs} 
\usepackage{hyperref}

\usepackage{amssymb}
\usepackage{amsmath}
\DeclareMathOperator*{\argmax}{arg\,max}
\DeclareMathOperator*{\argmin}{arg\,min}
\RequirePackage{algorithm}
\RequirePackage[noend]{algorithmic}

\usepackage{array}
\newcolumntype{L}[1]{>{\raggedright\let\newline\\\arraybackslash\hspace{0pt}}m{#1}}
\newcolumntype{C}[1]{>{\centering\let\newline\\\arraybackslash\hspace{0pt}}m{#1}}
\newcolumntype{R}[1]{>{\raggedleft\let\newline\\\arraybackslash\hspace{0pt}}m{#1}}

\usepackage[accepted]{icml2019}

\newcommand{\ourNAS}{IT-NAS}
\newcommand{\ourLCP}{IT-LCE}

\begin{document}

\twocolumn[
\icmltitle{Inductive Transfer for Neural Architecture Optimization}



\icmlsetsymbol{equal}{*}

\begin{icmlauthorlist}
\icmlauthor{Martin Wistuba}{ibm}
\icmlauthor{Tejaswini Pedapati}{ibm}
\end{icmlauthorlist}

\icmlaffiliation{ibm}{IBM Research}

\icmlcorrespondingauthor{Martin Wistuba}{martin.wistuba@ibm.com}
\icmlcorrespondingauthor{Tejaswini Pedapati}{tejaswinip@us.ibm.com}

\icmlkeywords{metalearning, transfer learning, deep learning, neural architecture selection, learning curve prediction, early stopping}

\vskip 0.3in
]



\printAffiliationsAndNotice{}  

\begin{abstract}
The recent advent of automated neural network architecture search led to several methods that outperform state-of-the-art human-designed architectures.
However, these approaches are computationally expensive, in extreme cases consuming GPU years.
We propose two novel methods which aim to expedite this optimization problem by transferring knowledge acquired from previous tasks to new ones.
First, we propose a novel neural architecture selection method which employs this knowledge to identify strong and weak characteristics of neural architectures across datasets.
Thus, these characteristics do not need to be rediscovered in every search, a strong weakness of current state-of-the-art searches.
Second, we propose a method for learning curve extrapolation to determine if a training process can be terminated early.
In contrast to existing work, we propose to learn from learning curves of architectures trained on other datasets to improve the prediction accuracy for novel datasets.
On five different image classification benchmarks, we empirically demonstrate that both of our orthogonal contributions independently lead to an acceleration, without any significant loss in accuracy.
\end{abstract}

\section{Introduction}
Deep learning techniques have been the key to major improvements in machine learning in various domains such as image and speech recognition and machine translation.
Traditionally, people built handcrafted neural network architectures for a particular task or they manually adapted popular networks, such as ResNet, DenseNet, Inception network, to the task at hand.
As finding an architecture manually is extremely arduous and requires an exploration of several network architectures, there has been an increased effort in automating it.
To this end, there have been several automated neural architecture search algorithms, mainly based on evolutionary algorithms or reinforcement learning \cite{Zoph2017,Real2017,Zoph2017a,Baker2017,Zhong2017,Liu2018,Pham2018}.
However, one challenge of these algorithm remains the computational effort to find the best network.
Most of these algorithms comprise of three components: the search space, the search strategy and the performance estimation of an architecture.
In our paper, we propose techniques to accelerate the search strategy and performance estimation.

Search Strategy:
We identified one common feature of all currently existing methods as a tremendous shortcoming: all methods search from scratch.
Therefore, we propose an approach which kick-starts the search based on inductive transfer.
We make the assumption that we have trained neural architectures already on other datasets and collected the metaknowledge of the experiments.
This includes information such as the topology and how the validation accuracy develops with growing number of epochs (learning curve).
However, it excludes the model parameters itself.
First, we use a predictive Bayesian model to extract a latent representation from the metaknowledge to transfer information about architectures across datasets.
Second, we use this model to guide the architecture selection on an unseen dataset.

Performance Estimation: To avoid training the network to completion to validate it, one can estimate the performance by leveraging learning curve extrapolation \cite{Domhan2015,Chandrashekaran2017,Klein2017,Baker2017a}. Terminating the training of poor performing networks early allows for a significant reduction in the computational effort.
However, some of these methods are based on prediction models which require a sufficient amount of training data \cite{Chandrashekaran2017,Klein2017,Baker2017a}.
Before applying them successfully, the data is generated by training architectures until completion.
Other approaches do not require training data but can only be applied after observing a sufficiently long learning curve \cite{Domhan2015}.
We address the issues of the existing methods with a new Bayesian transfer learning curve extrapolation method.
This method is able to utilize learning curves gathered across different datasets in order to improve its extrapolations for a new dataset.
As a consequence, we need only a select few learning curves from the current dataset to extrapolate learning curves accurately.

In conclusion, our contributions are
\begin{itemize}
 \item Formulation neural architecture selection and learning curve extrapolation as a transfer learning problem.
 \item Development of a learning curve extrapolation method to improve the accuracy of the forecasts.
 This is the \textit{very first} method to improve learning curve extrapolation by means of inductive transfer.
 Empirical evidence gathered on five different datasets indicates a speed-up by a factor of more than 16 when early stopping is employed, saving on average about 350 GPU hours.
 \item Derivation of a neural architecture selection method based on inductive transfer to accelerate the optimization.
 We conduct an elaborate comparison to competitor methods and human designed architectures on up to five different image classification benchmarks.
 State-of-the-art results are obtained on CIFAR-10 and CIFAR-100 in less than 8 hours of search time.
\end{itemize}

\begin{figure*}
\centering
\includegraphics[width=0.33\textwidth]{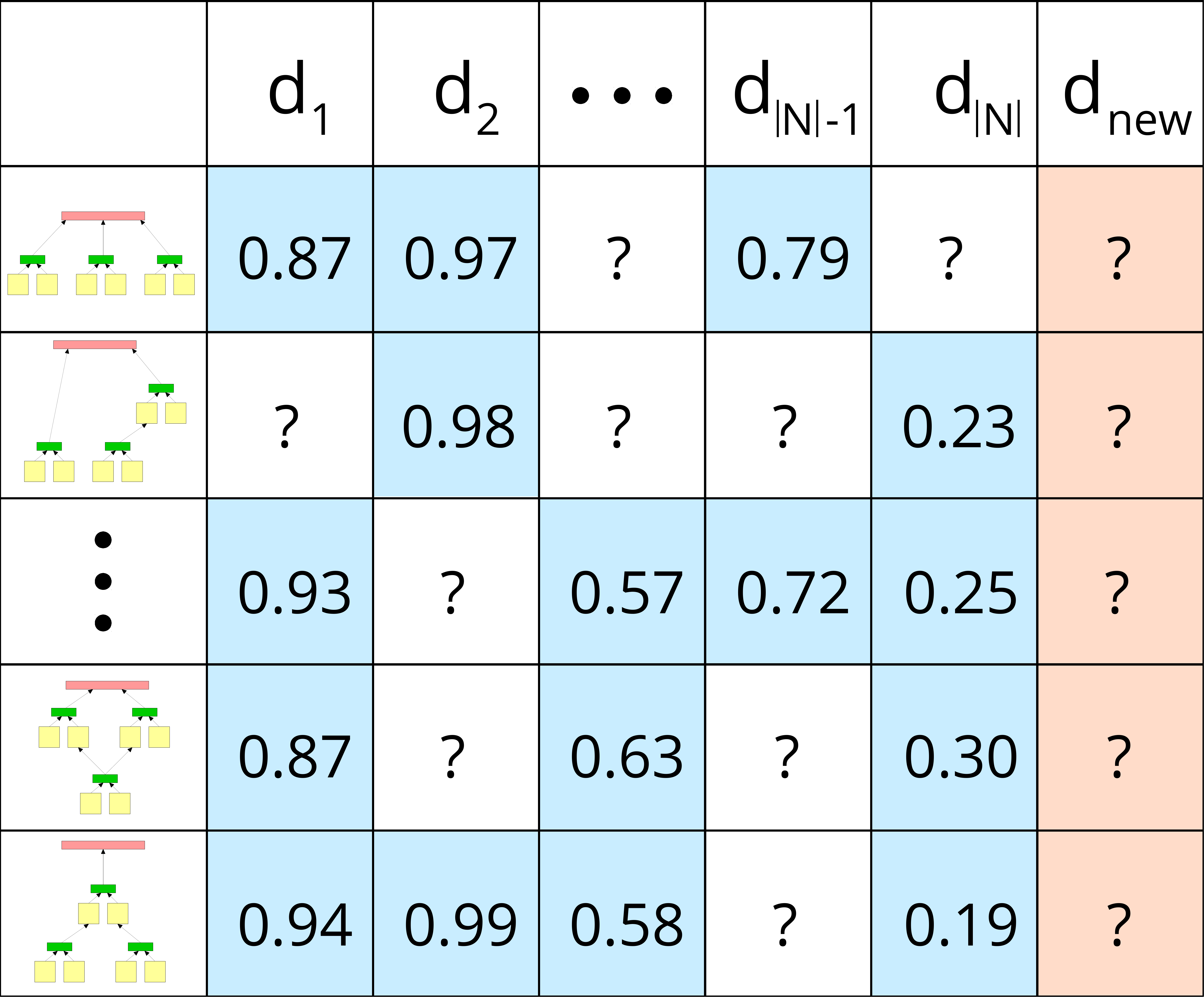}
\hspace{1cm}
\includegraphics[width=0.33\textwidth]{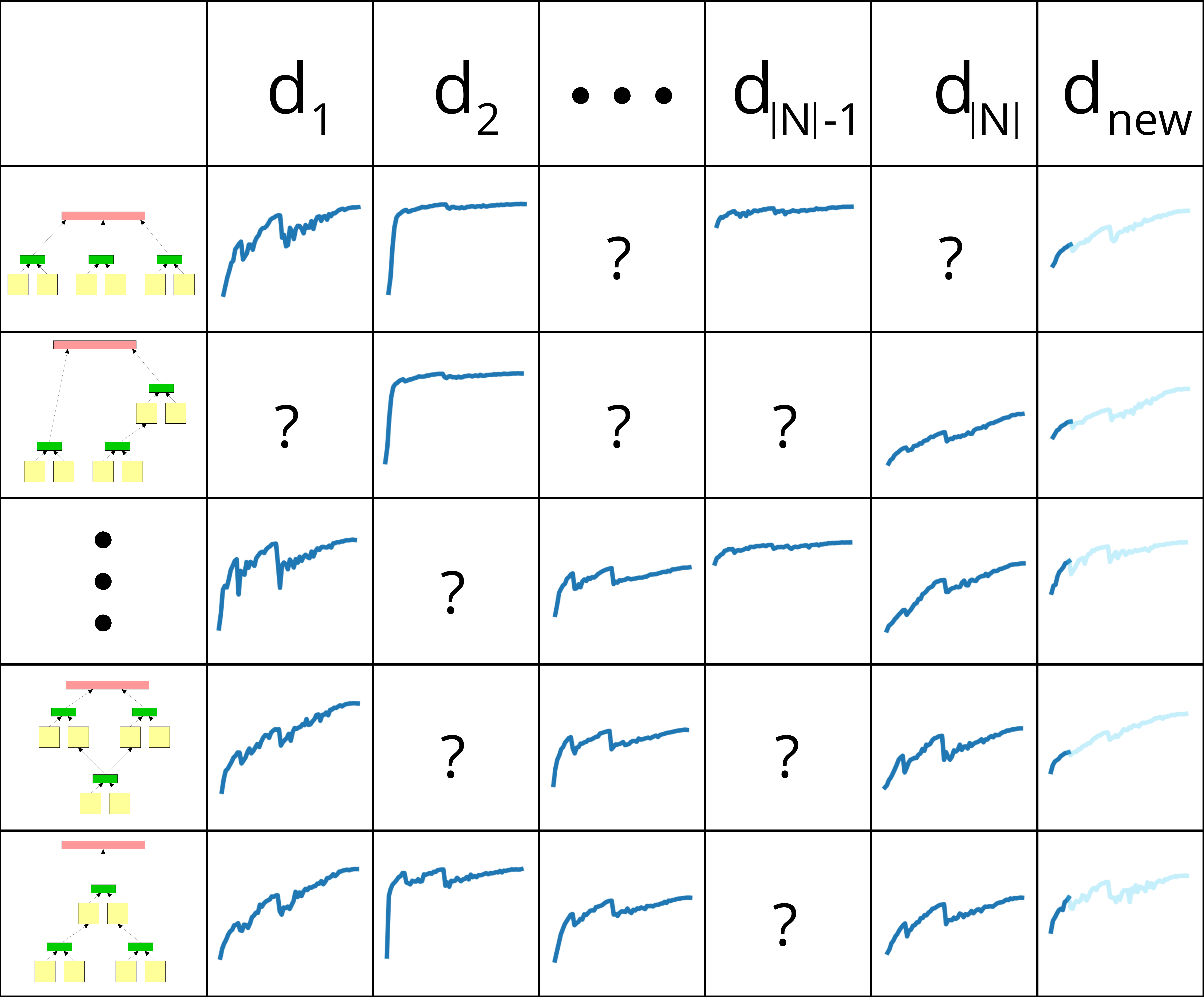}
\caption{The two problems addressed in this work using inductive transfer.
Left: Predicting the accuracy of a neural architecture based on observations on previous tasks.
Right: Extrapolating learning curves considering previous tasks and partially observed learning curves.}
\label{fig:problem-definition}
\end{figure*}

\section{Related Work}
\paragraph{Neural Architecture Search}
The increasing popularity of the automation of machine learning has led to several advances in automated neural architecture search.
Currently, it is dominated by methods based on evolutionary algorithms and reinforcement learning.

Neural Architecture Search (NAS) \cite{Zoph2017} is one of the early reinforcement learning methods achieving strong results on CIFAR-10 using 800 GPUs for an entire month.
Inspired by this, new methods were developed to reduce the search time by limiting the search space to repeating cells \cite{Zoph2017a,Zhong2017}, smart reusing \cite{Cai2017} or sharing of the model weights \cite{Pham2018,Liu2018b}.

Neuro-evolution is a well-studied method which started with evolving the network parameters \cite{Miller1989} and was later extended to evolve the architecture as well \cite{Stanley2002}.
\citet{Real2017} showed that neuro-evolution is able to find neural architectures with similar performance to human-designed architectures.
Very recently, the idea of learning repeating cells instead of the entire neural architecture network has been adopted for evolutionary algorithms \cite{Liu2018,Wistuba2018ECML}.
\citet{Miikkulainen2017} propose to co-evolve a set of cells and their connections.

The use of inductive transfer in the metalearning community is very common \cite{Vanschoren2018,Wistuba2018}.
However, so far these methods were developed to select traditional machine learning models (e.g. SVM, Random Forest) or tune their hyperparameters for tabular data \cite{Bardenet2013,Wistuba2015ICDM}.
Most of these methods also rely on metafeatures, i.e. features describing the properties of the dataset which are often not defined for unstructured data \cite{Yogatama2014,Schilling2016}.
Furthermore, many methods also expect that a possible candidate can be expressed with a fixed size and low-dimensional vector which is not the case for neural architectures \cite{Wistuba2015DSAA,Wistuba2016ECML,Schilling2015}.
For all those reasons most methods are not applicable to the problem of architecture selection.
Thus, we propose a novel method based on these principles to select neural architectures.

\paragraph{Learning Curve Prediction}
Most of the prior work for learning curve prediction is based on the idea of extrapolating the partial learning curve by using a combination of continuously increasing basic functions.
\citet{Domhan2015} define a set of 11 parametric basic functions, estimate their parameters and combine them in an ensemble.
\citet{Klein2017} propose a heteroscedastic Bayesian model which learns a weighted average of the basic functions.
\citet{Chandrashekaran2017} do not use basic functions but use previously observed learning curves for the current dataset.
An affine transformation for each previously seen learning curve is estimated by minimizing the mean squared error with respect to the partial learning curve.
The best fitting extrapolations are averaged as the final prediction.
\citet{Baker2017a} use a different procedure.
They use a support vector machine to predict the final accuracy based on features extracted from the learning curves, its gradients, and the neural architecture itself.

The predictor by \citet{Domhan2015} is able to forecast without seeing any learning curve before but requires observing more epochs for accurate predictions.
The model by \citet{Chandrashekaran2017} requires seeing few learning curves to extrapolate future learning curves.
However, accurate forecasts are already possible after few epochs.
Algorithms proposed by \citet{Klein2017} and \citet{Baker2017a} need to observe many full-length learning curves before providing any useful forecasts.
However, this is prohibiting in the scenarios where learning is time-consuming such as in large convolutional neural networks.

\section{Architecture Selection and Learning Curve Extrapolation with Inductive Transfer}

In this section we formulate neural architecture selection and learning curve extrapolation as a transfer learning task.
Furthermore, we propose two novel methods to solve these tasks accordingly.

\subsection{Inductive Transfer for Architecture Selection}

Our problem assumption is that we have access to previously trained neural network architectures for different datasets (see Figure \ref{fig:problem-definition} left).
Thus, we have a set of records $\mathcal{D}$ of which network architecture $n\in N$ was trained on dataset $d\in D$ with a validation accuracy $a_{n,d}$.
This set $\mathcal{D}$ represents the \textit{metaknowledge}.
The task is to predict $a_{n,d}$ for unseen tuples $\left(n,d\right)$ and in particular for datasets for which no observations are available in $\mathcal{D}$.
Our method is based on the biased matrix factorization model \cite{Koren2009}.
We can use it to predict the validation accuracy for a network architecture $n$ trained on a dataset $d$,
\begin{equation}
 \hat{a}_{n,d} = b_0 + b_n + b_d + v_n^T u_d\enspace.\label{eq:biased-mf}
\end{equation}
The model contains a global bias $b_{0}\in\mathbb{R}$, a network architecture and dataset bias $b_{n}, b_{d}\in\mathbb{R}$, and the latent network architecture/dataset representation $u_{n}, v_{d}\in\mathbb{R}^{f}$.
As we will see later in Section \ref{sub:nas}, the point estimate alone is not sufficient.
Instead, we take advantage of the posterior $p\left(a_{n,d}|\mathcal{D},\boldsymbol{\theta}\right)$.

\subsubsection{A Variational Prediction Model}\label{sub:opt-model}
We use variational learning to derive an equivalent Bayesian model with likelihood
\begin{equation}
 p\left(\mathcal{D}|\mathbf{w}\right) = \prod_{\left(n,d\right)\in\Omega}\mathcal{N}\left(a_{n,d}|b_0 + b_n + b_d + v_n^T u_d,\sigma^2\right).\label{eq:opt-model}
\end{equation}
where $\Omega\subseteq N\times D$ is the index set of observed network architecture evaluations and $\mathbf{w}$ all model parameters.
The idea of variational learning is to approximate intractable integrals by approximating posterior probabilities of unobserved variables.
This enables us to do statistical inference over these variables.
In our case, we are approximating $p\left(\mathbf{w}|\mathcal{D}\right)$ by a variational distribution $q\left(\mathbf{w}|\boldsymbol{\theta}\right)$.
We estimate the parameters $\boldsymbol{\theta}$ of the variational distribution by minimizing the Kullback-Leibler divergence of the exact Bayesian posterior on the weights from the variational approximation:
\begin{align}
 \boldsymbol{\theta}^{*} &= \argmin_{\boldsymbol{\theta}} D_{\mathrm{KL}}\left(q\left(\mathbf{w}|\boldsymbol{\theta}\right)\parallel p\left(\mathbf{w}|\mathcal{D}\right)\right)\\
 &= \argmin_{\boldsymbol{\theta}}\int q\left(\mathbf{w}|\boldsymbol{\theta}\right)\log\frac{q\left(\mathbf{w}|\boldsymbol{\theta}\right)}{p\left(\mathbf{w}\right)p\left(\mathcal{D}|\mathbf{w}\right)}\mathrm{d}\mathbf{w}\\
 &= \argmin_{\boldsymbol{\theta}} D_{\mathrm{KL}}\left(q\left(\mathbf{w}|\boldsymbol{\theta}\right)\parallel p\left(\mathbf{w}\right)\right)\nonumber\\
 &\phantom{=}\ - \mathbb{E}_{q\left(\mathbf{w}|\boldsymbol{\theta}\right)}\left[\log p\left(\mathcal{D}|\mathbf{w}\right)\right]\label{eq:objective-function}	\enspace.
\end{align}
This cost function is known as the evidence lower bound \cite{Saul1996} or the variational free energy \cite{Neal1993}.
It has two components.
The first term ensures that the variational distribution is similar to the prior over the parameters.
The second term is the fit to the data.
Thus, optimizing the cost function makes sure that the variational distribution satisfies both the complexity of the data $\mathcal{D}$ and our prior $p\left(\mathbf{w}\right)$.

The model parameters in Equation \eqref{eq:biased-mf} are $\mathbf{w}=\left(b_0, b_n, b_d, U, V\right)$.
We select independent Gaussian priors for them: $p\left(b_0\right)=\mathcal{N}\left(0,\sigma_{b_0}^2\right)$, $p\left(b_n\right),p\left(b_d\right)=\mathcal{N}\left(0,\sigma_{b}^2\right)$, $p\left(u_{d,k}\right),p\left(v_{n,k}\right)=\mathcal{N}\left(0,\sigma_{l}^2\right)$.
We choose a diagonal Gaussian distribution, $\mathcal{N}\left(\boldsymbol{\mu},\text{diag}\left(\boldsymbol{\sigma}\right)\right)$, for the variational posterior.
Thus, $\boldsymbol{\theta}=\left(\boldsymbol{\mu},\boldsymbol{\sigma}\right)$.
We estimate $\boldsymbol{\theta}$ by optimizing Equation \eqref{eq:objective-function} using stochastic gradient descent.

\subsubsection{Neural Architecture Selection}\label{sub:nas}
We derive a neural architecture selection method using inductive transfer by combining our proposed Bayesian model with Bayesian Optimization \cite{Mockus1978} as shown in Algorithm \ref{alg:Optimization}.
Given the new dataset $d$, we initially face the cold start problem \cite{Koren2009}.
The metaknowledge $\mathcal{D}$ does not contain any example for the new dataset and hence, the model cannot estimate any useful values for $b_d$ and $u_d$.
We overcome this by starting with the globally best network.
Then, we select a network architecture based on the expected utility.
We use the improvement to measure the utility  \cite{Mockus1978},
\begin{equation}
u\left(n,d\right)=\max\left\{a_{n,d}-a_{d}^{\text{max}},0\right\}\enspace,\label{eq:Definition-Improvement}
\end{equation}
where $a_{d}^{\text{max}}$ is the best known validation accuracy.
Then, the expected improvement is
\begin{align}
&\operatorname{EI}\left(\left(n,d\right),p\left(a_{n,d}|\mathcal{D},\boldsymbol{\theta}\right)\right):=\mathbb{E}\left[u\left(n,d\right)|\mathcal{D},\boldsymbol{\theta}\right]\\
&=\int_{a_{d}^{\text{max}}}^{\infty}\left(a_{n,d}-a_{d}^{\text{max}}\right)p\left(a_{n,d}|\mathcal{D},\boldsymbol{\theta}\right)\mathrm{d}a_{n,d}\enspace .
\end{align}
The expected improvement will be high for network architectures where the predictive posterior distribution assumes values higher than $a_{d}^{\text{max}}$.
It has a closed form solution if $p\left(a_{n,d}|\mathcal{D},\boldsymbol{\theta}\right)$ is Gaussian distributed \cite{Mockus1978}.
Algorithm \ref{alg:Optimization} now sequentially selects the architecture with highest expected improvement, trains and evaluates it and adds the additional observation to the metaknowledge $\mathcal{D}$.
We continue until some criterion is reached, e.g. a time constraint.
\begin{algorithm}
\caption{Neural Architecture Selection with Inductive Transfer}
\label{alg:Optimization}
\begin{algorithmic}[1]
\REQUIRE{Metaknowledge $\mathcal{D}$, new dataset $d$}.
\ENSURE{Best neural architecture found.}
\STATE Let $\bar{a}_n$ be the normalized mean accuracy of network $n$ across all tested datasets in $D$.
\STATE $n^{*}\leftarrow\argmax_{n\in N}\bar{a}_{n}$
\WHILE{not converged}
\STATE Train $n^{*}$ on $d$ and get validation accuracy $a_{n^{*},d}$.
\STATE $\mathcal{D}\leftarrow\mathcal{D}\cup\left\{a_{n^{*},d}\right\}$
\IF{$a_{n^{*},d}>a_{d}^{\text{max}}$}
        \STATE $n^{\text{max}}, a_{d}^{\text{max}}\leftarrow n^{*}, a_{n^{*},d}$
\ENDIF
\STATE Update the predictive posterior $p\left(a_{n,d}|\mathcal{D},\boldsymbol{\theta}\right)$.
\STATE $n^{*}\leftarrow\argmax_{n\in N}\operatorname{EI}\left(\left(n,d\right),p\left(a_{n,d}|\mathcal{D},\boldsymbol{\theta}\right)\right)$
\ENDWHILE
\RETURN{$n^{\text{max}}$}
\end{algorithmic}
\end{algorithm}

\subsection{Inductive Transfer for the Extrapolation of Learning Curves}
With the term \textit{learning curve} we describe the validation accuracy of an iterative algorithm, as a function of the number of epochs.
We denote the learning curve for a network $n$ trained on dataset $d$ for $\hat{T}$ epochs by $\mathbf{t}_{n,d}\in\mathbb{R}^{\hat{T}}$.
Now $\mathbf{t}_{n,d,T}$ is the observed validation accuracy after training for $T$ epochs.
Extrapolation of learning curves is the task to forecast the validation accuracy of a neural architecture after $\hat{T}$ epochs based on observing the first $T$ epochs only, where $T<\hat{T}$.
Similar to the previous task, we assume that we have collected metaknowledge $\mathcal{D}$ which contains learning curves $\mathbf{t}_{n,d}$ for all $(n,d)\in\Omega$, where $\Omega\subseteq N\times D$ is a subset of all network dataset pairs (see Figure \ref{fig:problem-definition} right).
Accurate learning curve extrapolations enable us to terminate training processes early without regret but at lower computational cost.

We extend the probabilistic model of Equation \eqref{eq:opt-model} by additionally considering the information about the partially observed learning curve $\mathbf{t}_{n,d,1:T}$ and define the likelihood for each instance $\left(n,d\right)\in\Omega$ as
\begin{equation}
 \mathcal{N}\left(\mathbf{t}_{n,d,\hat{T}}|b_0 + b_n + b_d + v_n^T u_d + r \max_{1\leq i\leq T}\mathbf{t}_{n,d,i},\sigma^2\right).\label{eq:learning-curve-model}
\end{equation}
To reduce the number of parameters we decide to use only the maximum validation accuracy observed in the learning curve instead of the entire learning curve.
This leads to only one additional parameter for which we define $p\left(r\right)=\mathcal{N}\left(0,\sigma_{b_r}^2\right)$ and $q\left(r|\mu_r,\sigma_r\right)=\mathcal{N}\left(r|\mu_r,\sigma_r\right)$.
Inference is done according to Section \ref{sub:opt-model}.

\subsubsection{The Early Termination Method}
We use our model in the early termination framework by \citet{Domhan2015} described in Algorithm \ref{alg:early-stopping}.
It determines whether training the neural architecture on a dataset $d$ should be stopped based on the partially observed learning curve $\mathbf{t}_{n,d}$.
If the best seen accuracy in the current run is higher than any accuracy seen on this dataset before ( $\mathbf{t}_{d}^{\text{max}}$), training is never stopped early (Line \ref{alg:train-to-end}).
Else, the probability of improvement \cite{Kushner1964} is computed (Line \ref{alg:compute-pi}).
Thus, we define the utility of a learning curve as
\begin{equation}
u\left(n,d\right)=\begin{cases}
1 & \text{if }\max_{1\leq i\leq T}\mathbf{t}_{n,d,i}>\mathbf{t}_{d}^{\text{max}}\\
0 & \text{otherwise}
\end{cases}
\end{equation}
and the probability of improvement derives as
\begin{align}
&\operatorname{PI}\left(\left(n,d\right),p\left(\mathbf{t}_{n,d,\hat{T}}|\mathcal{D},\boldsymbol{\theta}\right)\right) :=  \mathbb{E}\left[u\left(n,d\right)|\mathcal{D},\boldsymbol{\theta}\right]\\
& = \int_{\mathbf{t}_{d}^{\text{max}}}^{\infty} p\left(\mathbf{t}_{n,d,\hat{T}}|\mathcal{D},\boldsymbol{\theta}\right)\mathrm{d} \mathbf{t}_{n,d,\hat{T}}\enspace.
\end{align}
If the probability of improvement exceeds a threshold $\delta$, we continue training the architecture.
Otherwise, we terminate early.
\begin{algorithm}
\caption{Early Termination Method}
\label{alg:early-stopping}
\begin{algorithmic}[1]
\REQUIRE{Metaknowledge $\mathcal{D}$, new dataset $d$, neural architecture $n$}.
\ENSURE{Learning curve.}
\FOR{$T\leftarrow 1\ldots\hat{T}$}
\STATE Train $n$ on $d$ for one epoch and observe $\mathbf{t}_{n,d,T}$.
\IF{$\max_{1\leq i\leq T}\mathbf{t}_{n,d,i}>\mathbf{t}_{d}^{\text{max}}$}\label{alg:train-to-end}
  \STATE \textbf{continue}
\ELSIF{$\operatorname{PI}\left(\left(n,d\right),p\left(\mathbf{t}_{n,d,\hat{T}}|\mathcal{D},\boldsymbol{\theta}\right)\right) \leq \delta$}\label{alg:compute-pi}
  \RETURN{$\mathbf{t}_{n,d}$}
\ENDIF
\ENDFOR
\RETURN{$\mathbf{t}_{n,d}$}
\end{algorithmic}
\end{algorithm}

\section{Experimental Evaluation}
We will compare our models to a representative set of competitor methods for the two challenging tasks, neural architecture optimization and learning curve extrapolation.
However, at first we introduce the used datasets, the defined search space and how we obtained the metaknowledge for our experiments.

\begin{figure*}
\centering
\includegraphics[width=0.33\textwidth]{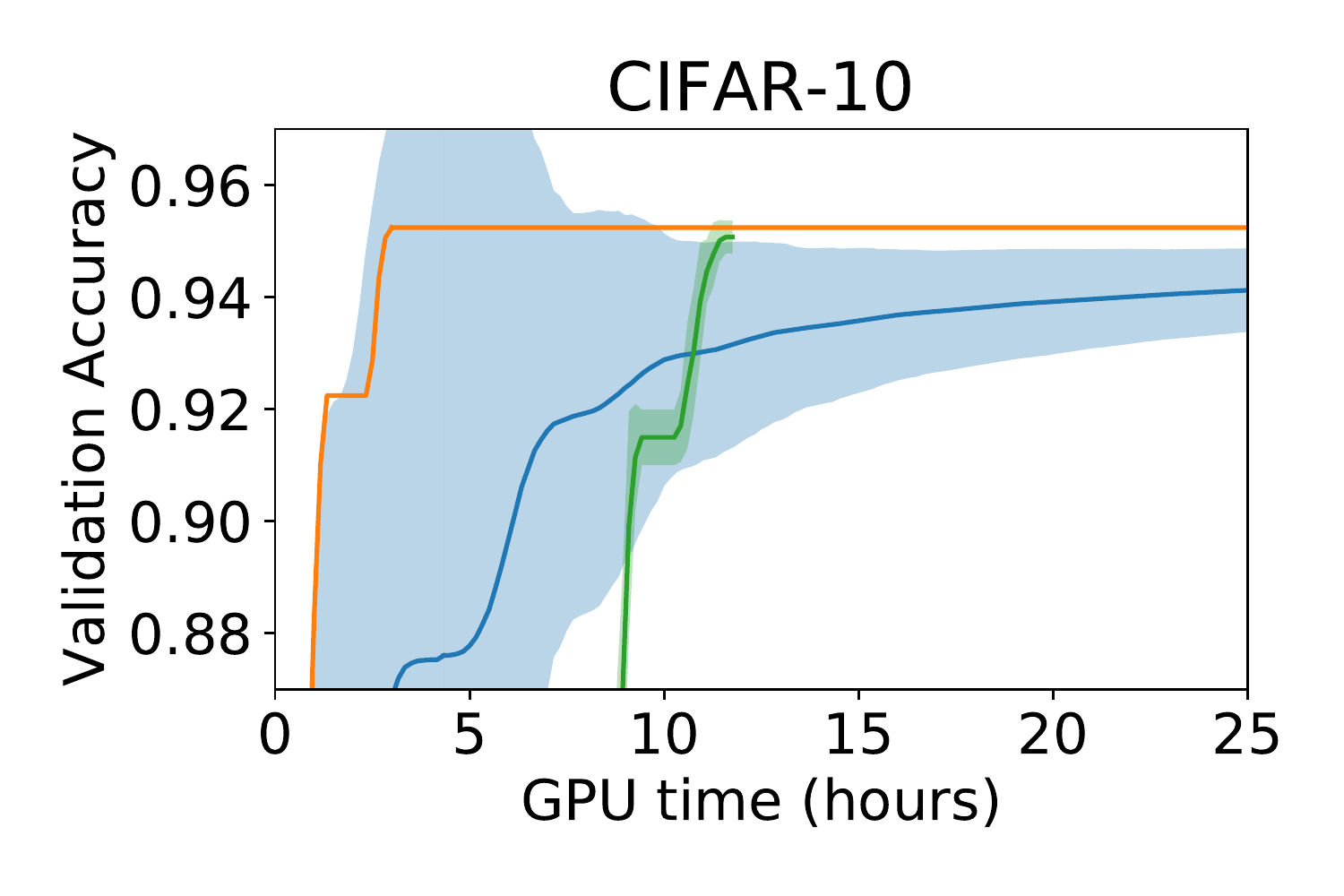}
\includegraphics[width=0.33\textwidth]{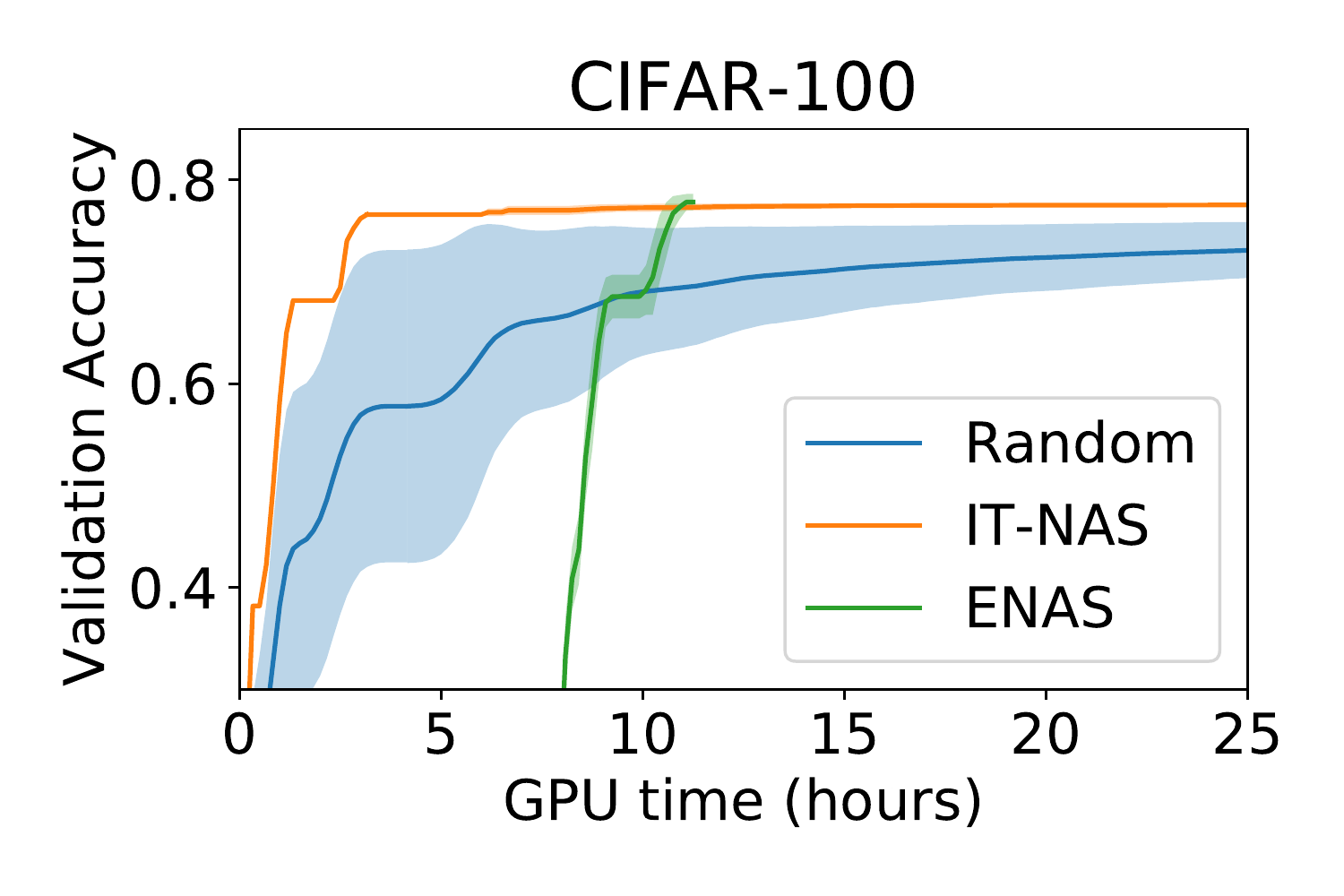}
\includegraphics[width=0.33\textwidth]{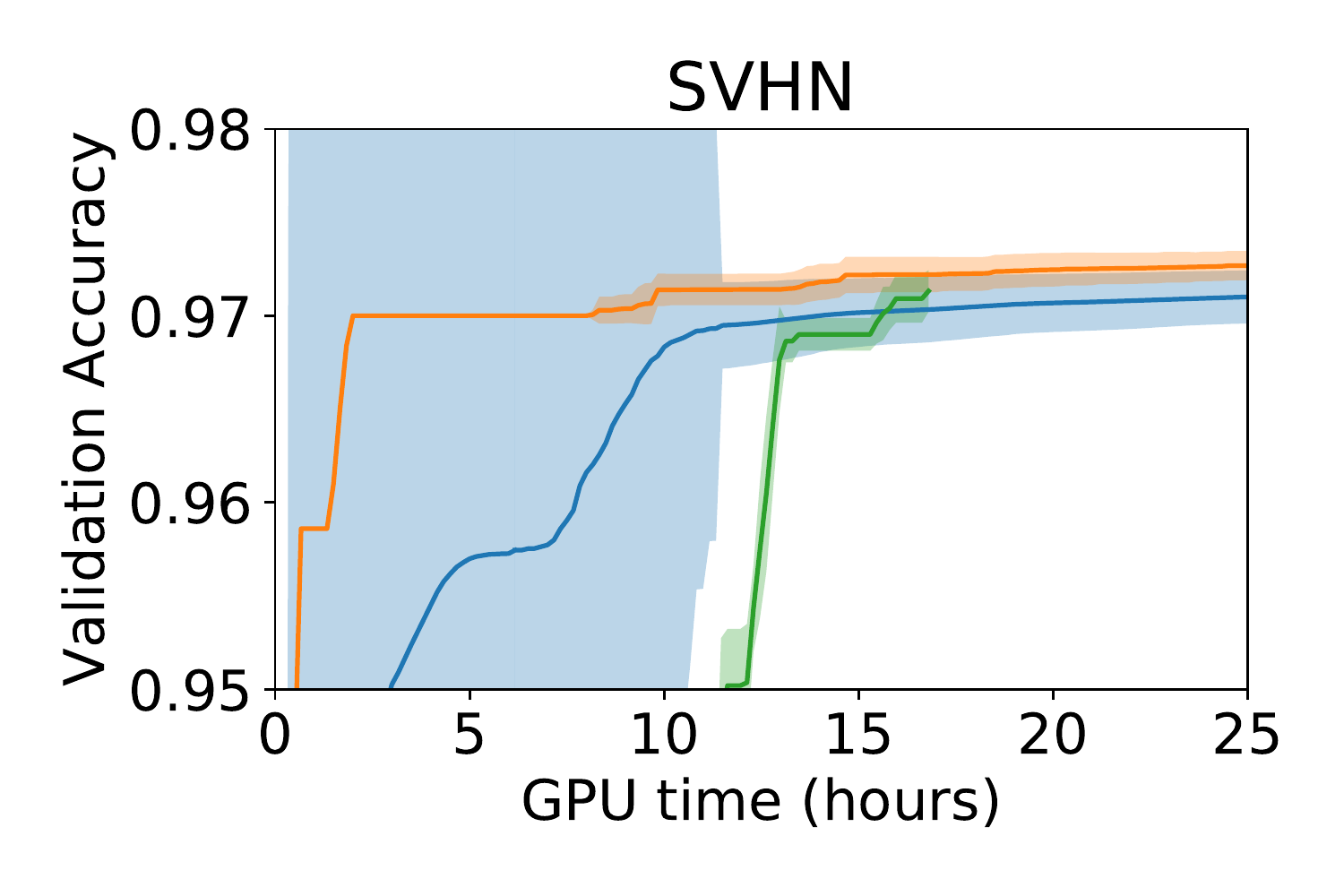}
\includegraphics[width=0.33\textwidth]{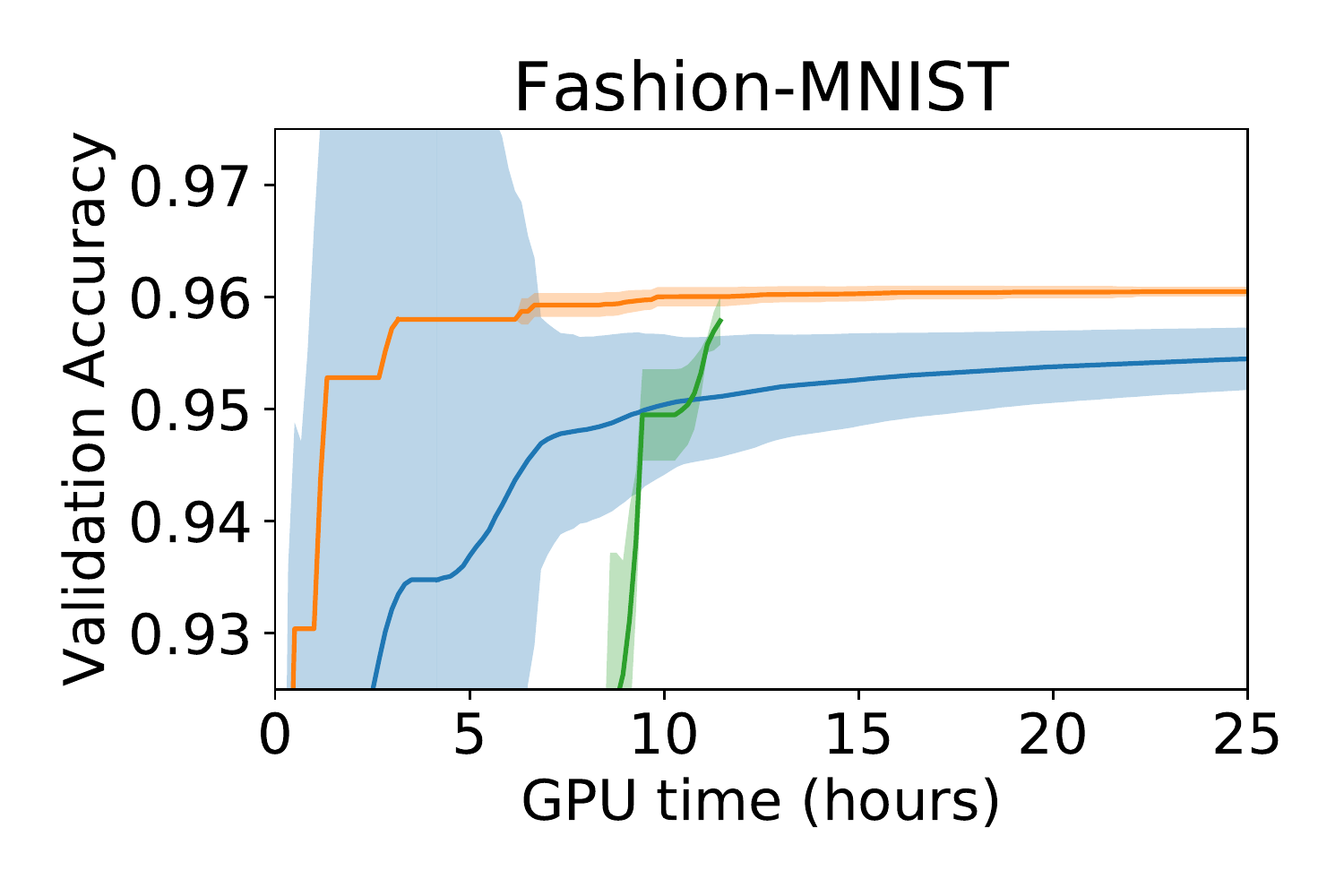}
\includegraphics[width=0.33\textwidth]{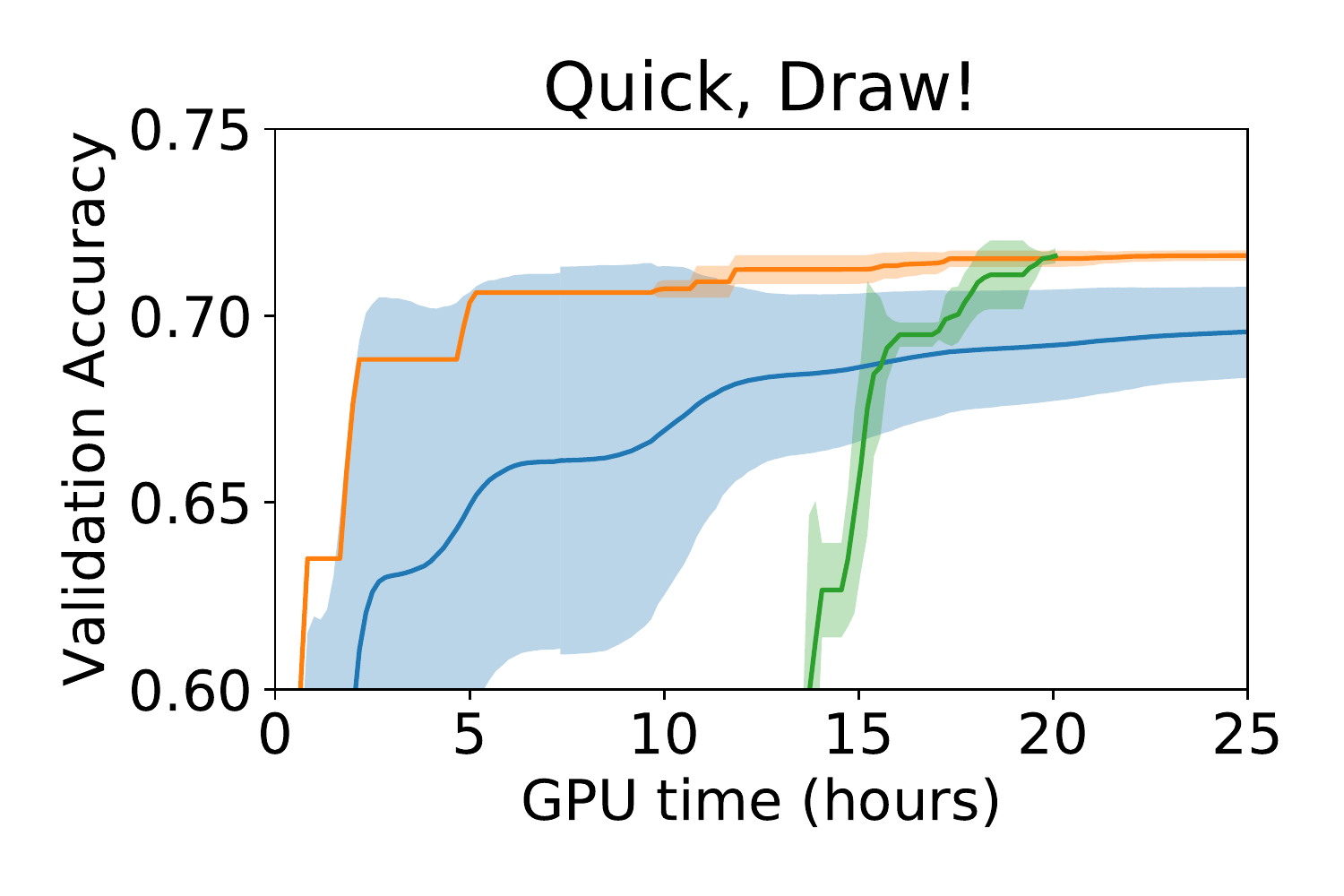}
\caption{\ourNAS{} achieves very good results early on and outperforms random search at any time.
ENAS provides similarly good results, however, at higher computational costs, in particular for larger datasets, e.g. ``Quick, Draw!''.}
\label{fig:opt-results-short}
\end{figure*}
\subsection{Datasets}
We evaluate our proposed methods on five different image classification datasets.
The CIFAR-10 dataset \cite{Krizhevsky2009} consists of colored 50,000 train and 10,000 test images of dimension $32\times 32$ where one needs to distinguish between ten classes.
CIFAR-100 \cite{Krizhevsky2009} has the same dimensions but one hundred instead of ten classes.
The Street View House Number (SVHN) \cite{Netzer2011} dataset is another ten class problem with images of size $32\times 32$.
It has about 73,257 training and 26,032 test images.
It contains about half a million additional images which we do not consider in our experiments.
Fashion-MNIST \cite{Xiao2017} is a more recent dataset intended to replace MNIST.
Like MNIST, it is a ten class problem with $28\times 28$ gray-scale images.
``Quick, Draw!'' is a dataset with about 50 million sketches with 345 classes created by over 15 million humans.
For computational reasons we randomly select a subset, 300 examples per class for train and 100 per class for test.
For each dataset we use a random subset of size 5,000 of the training set to validate a neural network architecture.

We use the same data preprocessing and data augmentation as also used by \cite{Pham2018}.
We subtract the channel mean from the image and divide it by the channel standard deviation.
We enlarge images by padding a margin of four pixels and randomly crop it back to the original dimension.
For all datasets but SVHN we apply random horizontal flipping.
Additionally, we use CutOut \cite{DeVries2017}.
By using the openly available code of \cite{Pham2018} to train an architecture, we ensure a fair comparison with ENAS.

\subsection{Search Space and Metaknowledge}\label{sub:search-space}
The micro search space is a very common architecture search space \cite{Zoph2017a,Pham2018}.
Its idea is to design convolution and reduction cells instead of the entire neural architecture.
To generate the data for our metaknowledge, we select 400 neural architectures at random from this search space.
All 400 architectures are trained on the five datasets described in the previous section and serve as our search space.
The architectures are trained with a Nesterov momentum and stochastic gradient descent with a cosine schedule with $l_{\text{min}}=0.001$, $l_{\text{max}}=0.05$, $T_0=10$ and $T_\text{mul}=2$ for 70 epochs \cite{Loshchilov2016}.
The $l_2$ weight decay is fixed to $10^{-4}$.
The metaknowledge in the following experiments contains all instances but those belonging to the current test dataset.

\subsection{Automated Architecture Optimization}
In this section, we discuss two different experiments.
In the first one we train each neural architecture for 70 epochs, resulting in lower accuracies.
However, it enables us to conduct more repetitions and hence to report reliable standard deviations.
In the second one we train for 630 epochs following \citet{Pham2018}, and compare with the competitor methods.

At first, we compare our proposed method \ourNAS{} to ENAS \cite{Pham2018} and random search.
Limiting random search and \ourNAS{} to the 400 randomly selected architectures, we can report results for 10,000 and 100 repetitions, respectively.
We then report the mean accuracy obtained by ENAS over five repetitions using the authors' implementation in Figure \ref{fig:opt-results-short}.
Across all datasets we notice very good results for \ourNAS{} within few hours, considerably outperforming ENAS and random search.
After this initial phase, \ourNAS{} shows small but steady improvement of up to 1\%.
ENAS achieves similarly good results but has drawbacks for easier datasets and those with more computational effort per epoch.
For example, SVHN is easier to solve such that a random search finds an equivalent neural architecture to ENAS within shorter time.
On the largest dataset, ``Quick, Draw!'', ENAS requires about 20 hours (almost twice as long as on CIFAR-10) before having reached the performance of \ourNAS{}.
In contrast, \ourNAS{} and random search trained few models before, enabling shorter search times if required.
Finally, both \ourNAS{} and random search can profit from early termination using learning curve extrapolation (as discussed in Section \ref{sub:experiments-lcp}) while ENAS cannot.

\begin{figure*}
\centering
\includegraphics[width=0.65\textwidth]{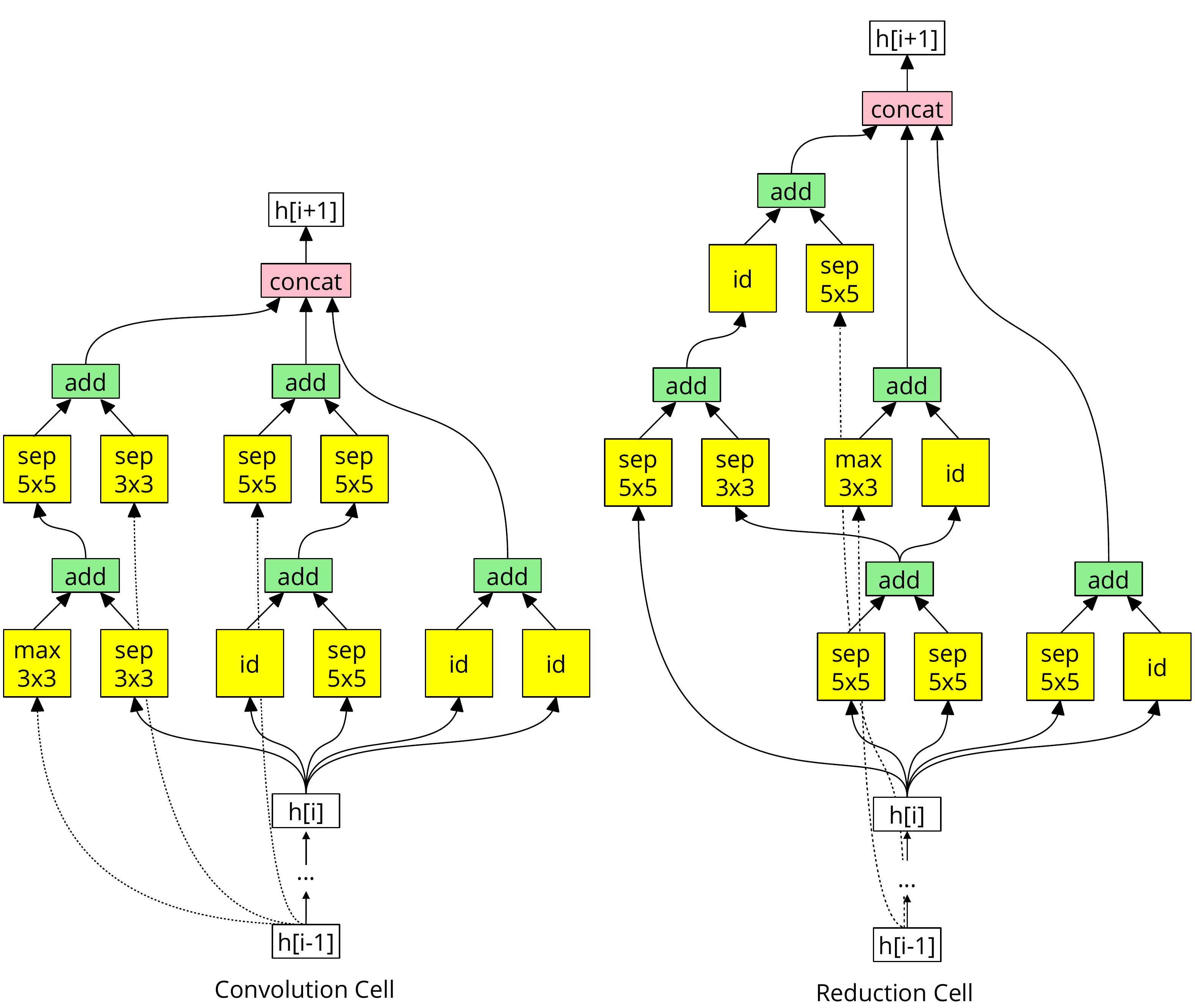}
\caption{Cells selected for CIFAR-10.}
\label{fig:cifar-10-cells}
\end{figure*}

\begin{figure*}
\centering
\includegraphics[width=0.99\textwidth]{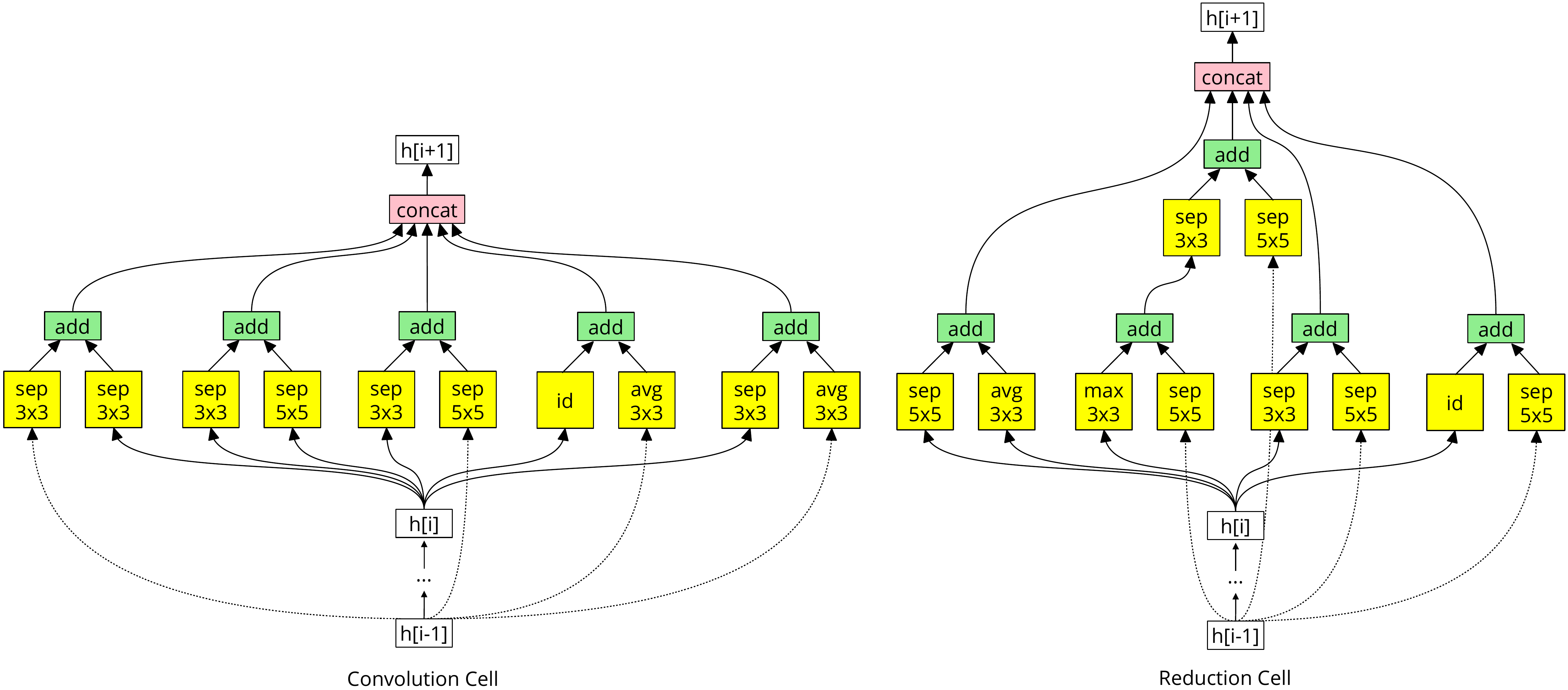}
\caption{Cells selected for CIFAR-100.}
\label{fig:cifar-100-cells}
\end{figure*}

Second, we compare our method against the competitor methods.
For this, we continue training the best neural architectures found in the previous experiment and report the corresponding errors in Table \ref{tab:opt-results}.
We repeated the experiments of ENAS on CIFAR-10 with the authors' code and achieve comparable results.
We report a shorter run time because we use an NVIDIA Tesla V100 while the authors used the slower NVIDIA GTX 1080Ti GPU.
The impact of the metaknowledge is significant, enabling \ourNAS{} to outperform all competitors on CIFAR-10 and CIFAR-100 at no search costs and with a moderate number of model parameters.
Only ENAS reports better number for CIFAR-100.
However, using the same search budget, \ourNAS{} is also able to find a better architecture than ENAS.
These results can be potentially be improved by using the early stopping technique discussed in the next section.
We report the cells selected for the two benchmark datasets CIFAR-10 and CIFAR-100 in Figure \ref{fig:cifar-10-cells} and \ref{fig:cifar-100-cells}, respectively.

\begin{table*}[t]
\centering
\caption{Results on CIFAR-10 for various mobile architectures. Search time in GPU days.\label{tab:results-mobile}}
\centering
\begin{tabular}{lrrrrr}
\hline\noalign{\smallskip}
Method & Search Time & Error & Params\\
\noalign{\smallskip}
\hline
\noalign{\smallskip}
DenseNet-BC ($k=12$) \cite{Huang2017} & N/A & 4.51 & 0.8M \\
CondenseNet-86 \cite{Huang2017a} & N/A & 5.00 & 0.52M \\
\noalign{\smallskip}
\hline
\noalign{\smallskip}
NSGA-NET \cite{Lu2018} & 8 & 3.85 & 3.3M\\
NASNet-B \cite{Zoph2017a} & 2000 & 3.73 & 2.6M\\
RENA \cite{Zhou2018} & ? & 3.98 & 2.2M\\
Lemonade \cite{Elsken2018} & 56 & 4.6 & 0.88M\\
DPP-Net \cite{Dong2018} & 8 & 4.62 & 0.52M\\
\noalign{\smallskip}
\hline
\noalign{\smallskip}
IT-NAS-1 (mobile) & 0.00 & 4.06 & 0.57M\\
IT-NAS-2 (mobile) & 0.08 & 3.65 & 0.66M\\
\hline
\end{tabular}
\end{table*}

\begin{figure*}[t]
\centering
\includegraphics[width=0.99\textwidth]{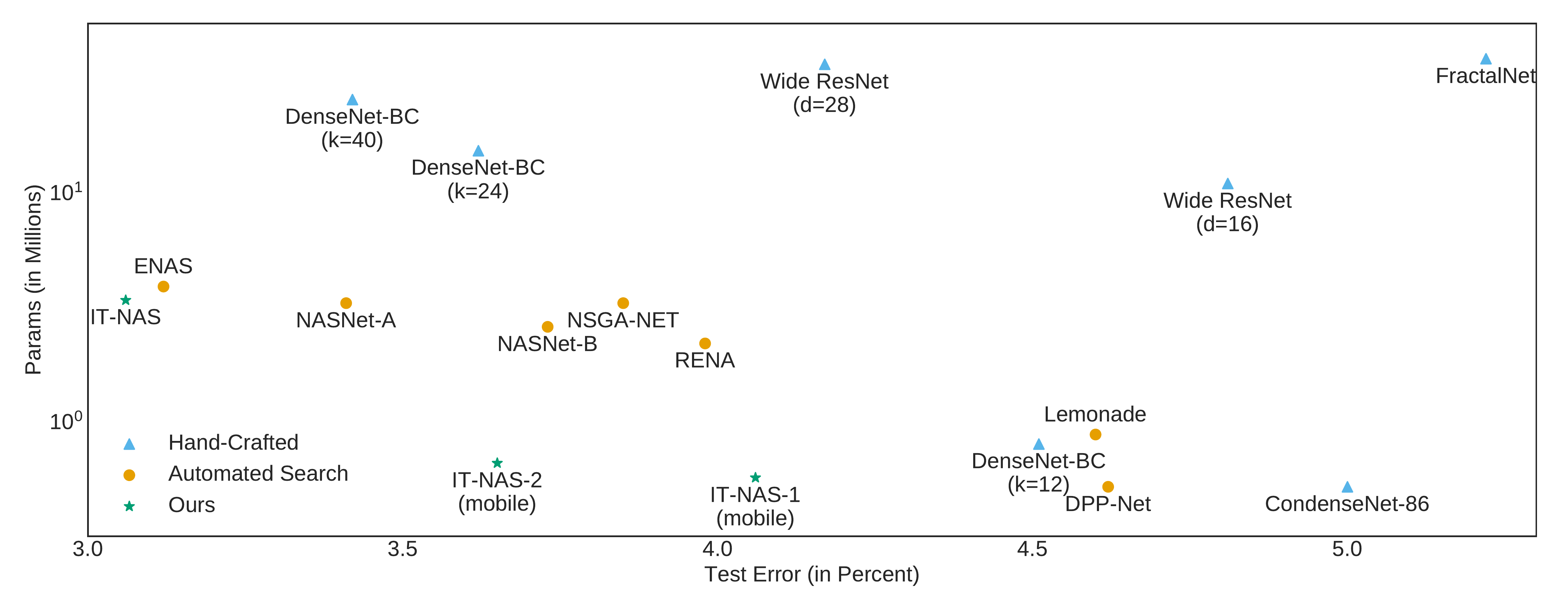}
\caption{Test error vs. number of parameters for various architectures.}
\label{fig:architecture plot}
\end{figure*}

\subsection{Mobile Architecture Selection}
Recently, more and more automated methods are described in order to take hardware constraints into account.
In a little experiment we want to show that our method can be applied without any big changes to achieve the same.
Our only change is the search space which is now limited to approximately 0.3 to 0.8 million parameters, our definition of a mobile architecture on CIFAR-10.
We report our results in Table \ref{tab:results-mobile}.
We notice that our method is orders of magnitudes faster than all competitor methods and provides the smallest error with very few parameters.
Our search algorithm is faster than before.
The reason for this is that mobile architectures are trained faster, so we only require at most few hours.

\subsection{Detailed Overview of Competitor Architectures}

We provide a detailed overview of the trade-off between number of parameters and classification error on CIFAR-10 in Figure \ref{fig:architecture plot}.
The architectures found by our method IT-NAS are well performing with respect to both metrics and thus are in the lower left corner of the plot.
They dominate most of the other proposed architectures and are not dominated by any other architecture.


\subsection{Learning Curve Extrapolation}\label{sub:experiments-lcp}
We conduct two different experiments to evaluate the quality of different learning curve extrapolation methods.
First, we compare the methods with respect to a rank correlation metric, indicating whether the predicted accuracy reflects the true ranking of the neural architectures.
Second, we use the exploration of learning curves to expedite a random search.
We use the authors' implementation of \citet{Domhan2015} and \citet{Chandrashekaran2017} to run these methods.

\subsubsection{Rank Correlation}
\begin{figure*}
\centering
\includegraphics[width=0.33\textwidth]{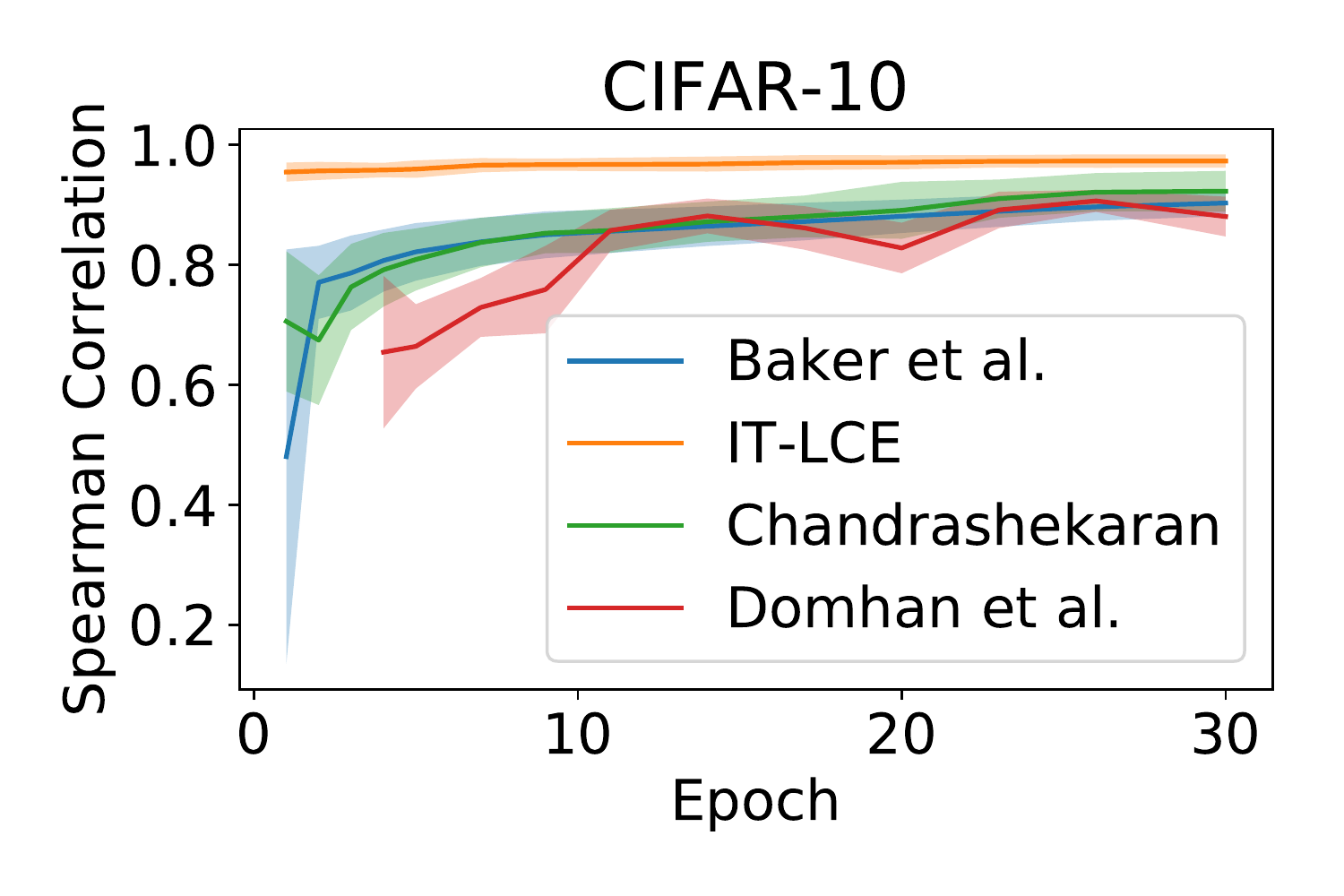}
\includegraphics[width=0.33\textwidth]{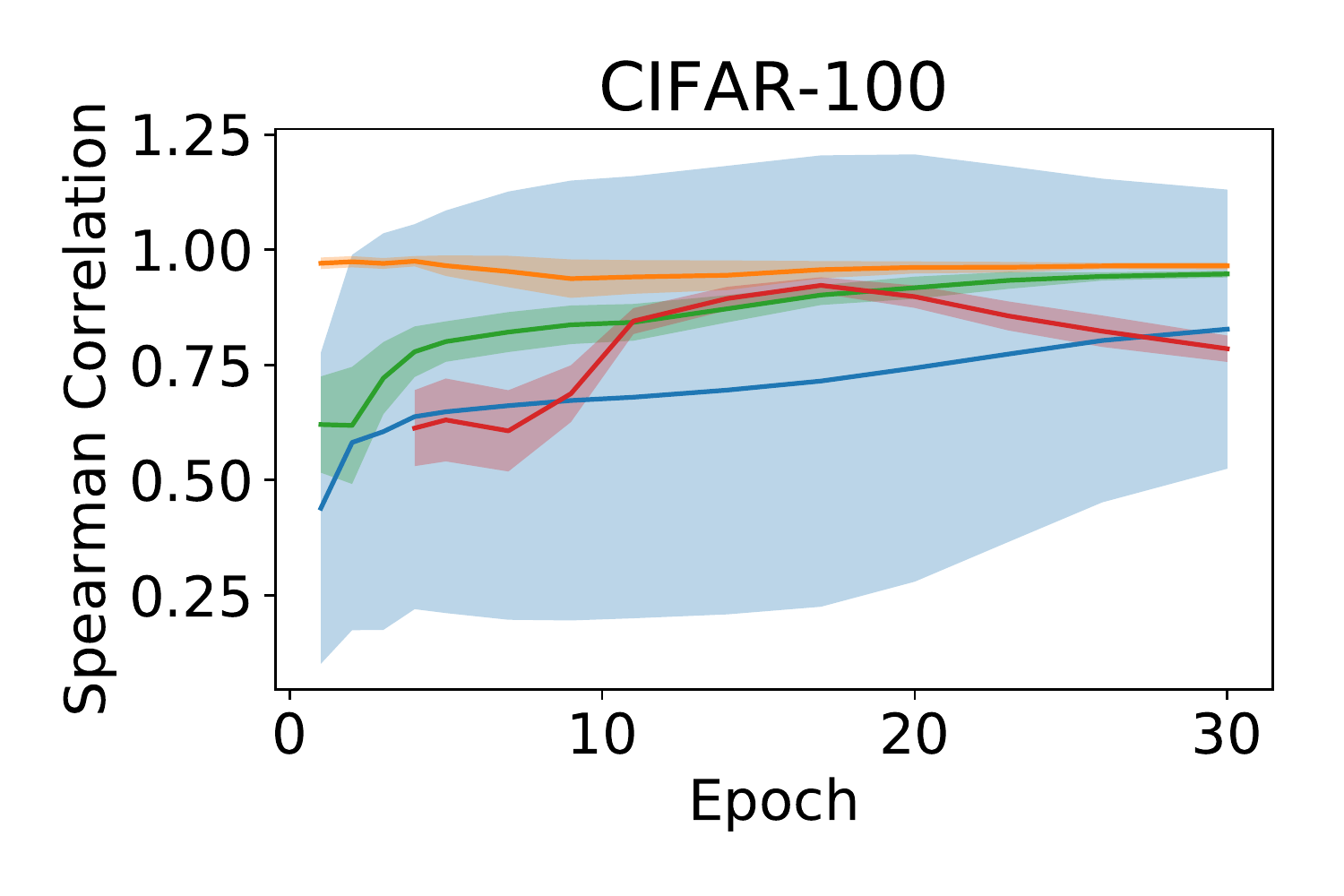}
\includegraphics[width=0.33\textwidth]{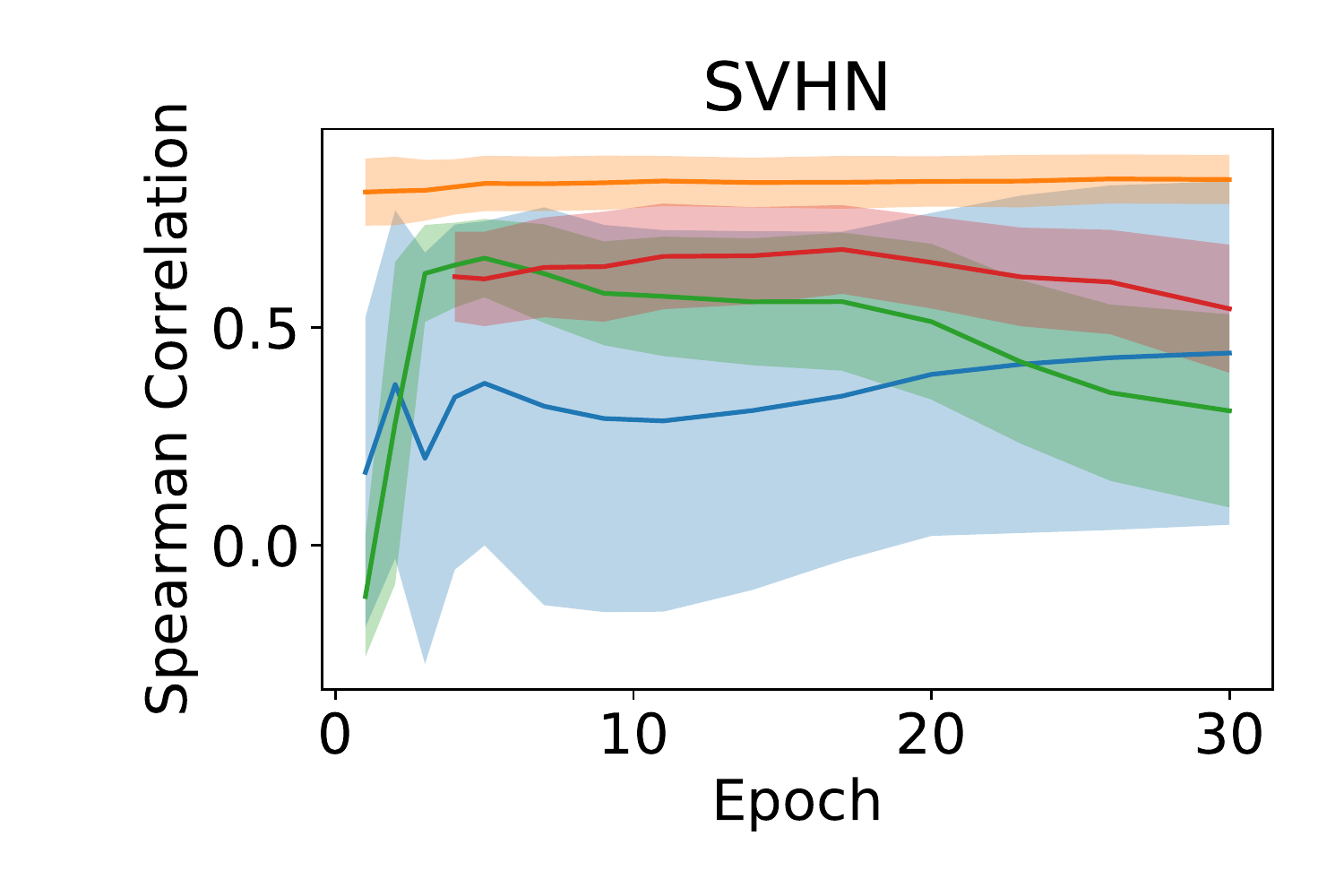}
\includegraphics[width=0.33\textwidth]{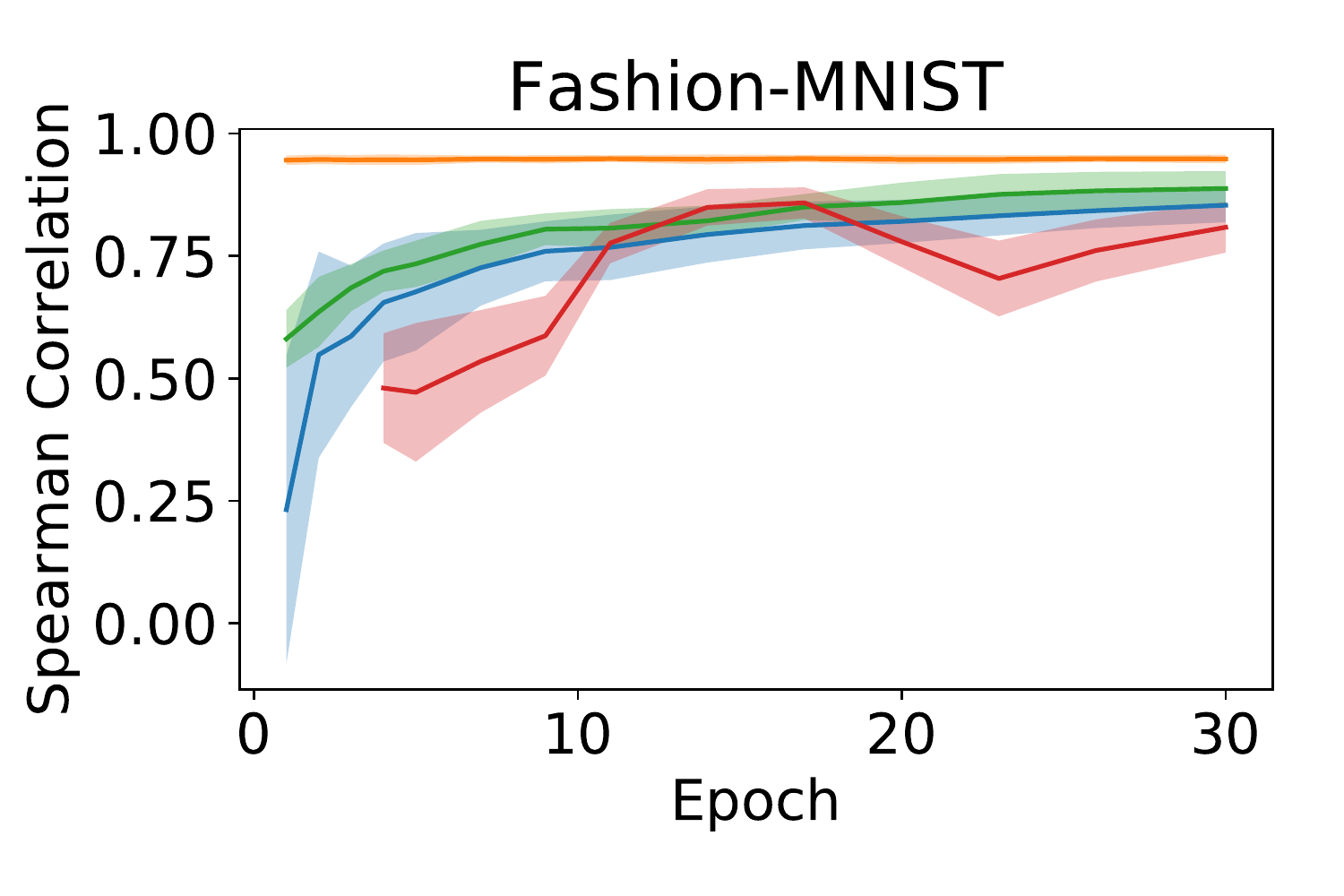}
\includegraphics[width=0.33\textwidth]{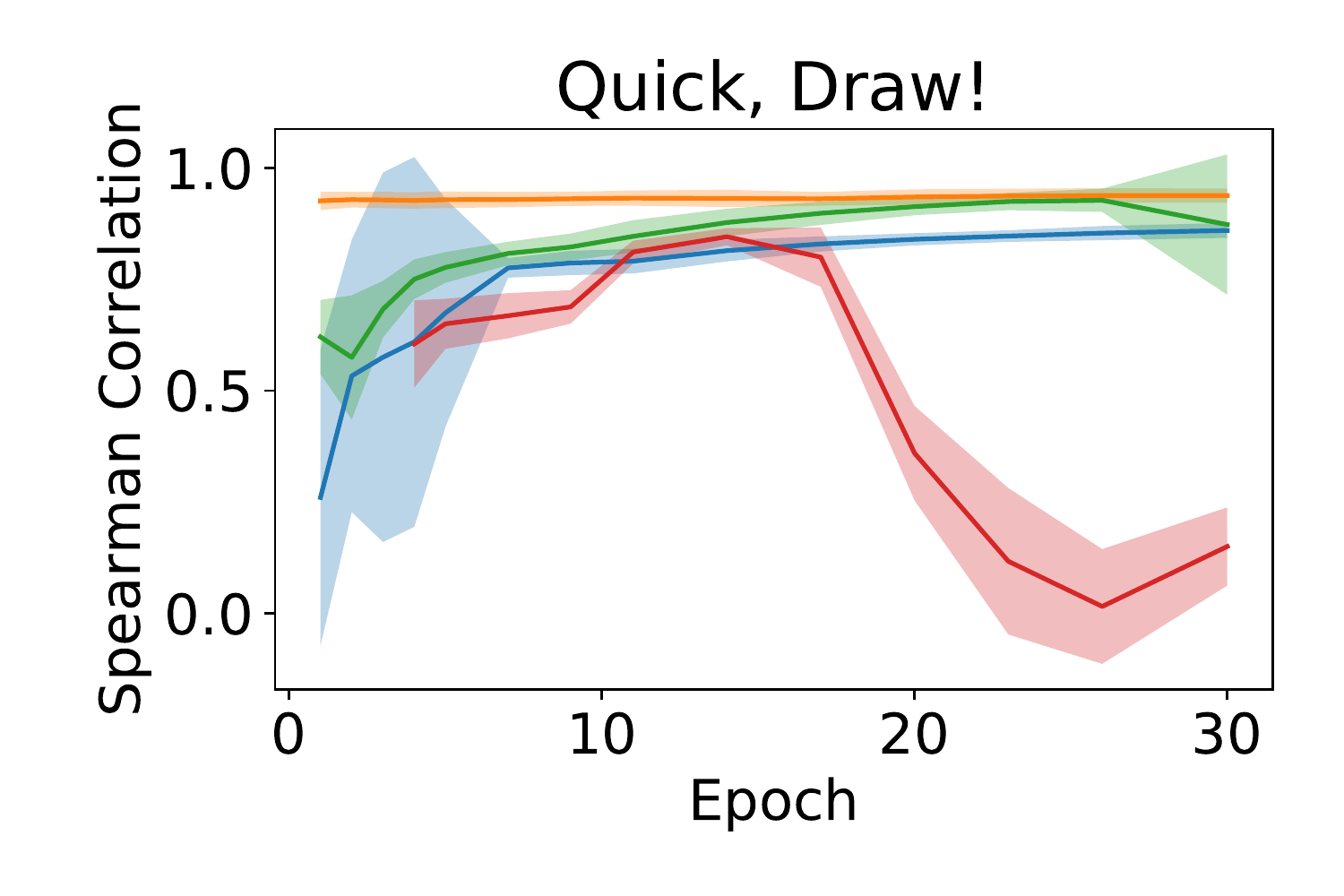}
\caption{The x-axis indicates the number of epochs observed of the learning curve to be extrapolated.
We report the mean of ten repetitions.
The shaded area is the standard deviation.
Our method \ourLCP{} outperforms its competitors on all datasets.}
\label{fig:lcp-rank-results}
\end{figure*}

In our first experiment we measure the rank correlation between the true accuracy of an architecture and the predictions of the four learning curve predictors.
We select five neural architectures from the search space described in Section~\ref{sub:search-space} at random.
The corresponding full learning curves are provided to the learning curve predictors which need to provide forecasts for the remaining architectures.
Additionally, we provide partial learning curves of varying length.
The method by \citet{Domhan2015} requires to see at least four epochs for a new learning curve in order to extrapolate it.
Thus, its results start at a later stage.
In Figure \ref{fig:lcp-rank-results} we report the results over ten repetitions.
Clearly, our proposed method using inductive transfer (\ourLCP{}) is outperforming its competitor methods across all five datasets with respect to the Spearman rank correlation coefficient.
At first, one would expect a growing rank correlation coefficient when more information about a learning curve is available (more epochs).
However, this does not turn out to be true for all methods and all datasets.
SVHN is a dataset where we observe this behavior for many methods.
The explanation is that SVHN is a more difficult dataset for this particular task.
The accuracy for different architectures is very close, making it harder to get the ranking (not the predicted accuracy!) of the neural architectures right.
This also explains the overall worse Spearman correlation for this dataset.
Hardly noticeable, \ourLCP{}'s rank correlation shows an increasing trend with growing number of epochs.

\begin{table*}
\caption{Classification error for CIFAR-10 and CIFAR-100.
The first block contains handcrafted neural architectures, the second block automated methods based on reinforcement learning and, the third, those based on neuro-evolution.
\ourNAS{} is outperforming all but one method for both datasets with a reasonable amount of model parameters.
Search duration in days. ENAS/\ourNAS{} architectures are trained for 630 epochs which leads to a smaller error compared to Figure \ref{fig:opt-results-short}. Block diagrams of the architectures can be found in Figure \ref{fig:cifar-10-cells} and \ref{fig:cifar-100-cells}.}
\label{tab:opt-results}
\centering
\begin{tabular}{p{0.5\textwidth}R{0.1\textwidth}R{0.06\textwidth}R{0.07\textwidth}R{0.06\textwidth}R{0.07\textwidth}}
\hline\noalign{\smallskip}
Method & Search & \multicolumn{2}{c}{CIFAR-10} & \multicolumn{2}{c}{CIFAR-100} \\
 & Duration  & Error & Params & Error & Params\\
\noalign{\smallskip}
\hline
\noalign{\smallskip}
ResNet \cite{He2016} reported by \cite{Huang2016}&  N/A & 6.41 & 1.7M &  27.22 & 1.7M \\
FractalNet \cite{Larsson2017} &  N/A & 5.22 & 38.6M &  23.30 & 38.6M \\
Wide ResNet ($d=16$) \cite{Zagoruyko2016} &  N/A & 4.81 & 11.0M &  22.07 & 11.0M \\
Wide ResNet ($d=28$) \cite{Zagoruyko2016} &  N/A & 4.17 & 36.5M &  20.50 & 36.5M \\
DenseNet-BC ($k=12$) \cite{Huang2017} &  N/A & 4.51 & \textbf{0.8M} &  22.27 & 0.8M \\
DenseNet-BC ($k=24$) \cite{Huang2017} &  N/A & 3.62 & 15.3M &  17.60 & 15.3M \\
DenseNet-BC ($k=40$) \cite{Huang2017} &  N/A & \textbf{3.42} & 25.6M &  17.18 & 25.6M \\
\noalign{\smallskip}
\hline
\noalign{\smallskip}
NAS no stride/pooling \cite{Zoph2017} &  22,400 & 5.50 & 4.2M &  - & -\\
NAS predicting strides \cite{Zoph2017} &  22,400 & 6.01 & \textbf{2.5M} &  - & -\\
NAS max pooling \cite{Zoph2017} &  22,400 & 4.47 & 7.1M &  - & -\\
NAS max pooling + more filters \cite{Zoph2017} &  22,400 & 3.65 & 37.4M &  - & -\\
NASNet \cite{Zoph2017a} &  2,000 & 3.41 & 3.3M &  - & -\\
MetaQNN \cite{Baker2017} &  100 & 6.92 & 11.2M &  27.14 & 11.2M\\
BlockQNN \cite{Zhong2017} &  96 & 3.6 & ? &  18.64 & ?\\
Efficient Architecture Search \cite{Cai2017} &  10 & 4.23 & 23.4M &  - & -\\
Efficient NAS \cite{Pham2018} &  0.45 & \textbf{2.89} & 4.6M &  - & -\\
Efficient NAS (ours) &  0.32 & 3.12 & 3.9M &  16.55 & 4.2M\\
\noalign{\smallskip}
\hline
\noalign{\smallskip}
Large-Scale Evolution \cite{Real2017} &  2,600 & 5.4 & 5.4M &  23.0 & 40.4M\\
Hierarchical Evolution \cite{Liu2018} &  300 & \textbf{3.75} & 15.7M &  - & -\\
CGP-CNN (ResSet) \cite{Suganuma2017} &  27.4 & 6.05 & \textbf{2.6M} &  - & -\\
CoDeepNEAT \cite{Miikkulainen2017} &  ? & 7.30 & ? &  - & -\\
\noalign{\smallskip}
\hline
\noalign{\smallskip}
\ourNAS{} (single shot) &  0 & 3.06 & 3.4M &  17.13 & 3.4M\\
\ourNAS{} &  0.32 & 3.06 & 3.4M &  16.38 & 4.2M\\
\hline
\end{tabular}
\end{table*}
\begin{table*}
\caption{Our method achieves the highest acceleration among all considered methods. The time required is reduced by a factor of 16 to 23. Time in hours, regret relative to the optimum of ``No Stopping''.}
\label{tab:random-search}
\begin{tabular}{p{0.19\textwidth}R{0.06\textwidth}R{0.05\textwidth}R{0.06\textwidth}R{0.05\textwidth}R{0.06\textwidth}R{0.05\textwidth}R{0.06\textwidth}R{0.05\textwidth}R{0.06\textwidth}R{0.05\textwidth}}
\hline\noalign{\smallskip}
Method & \multicolumn{2}{c}{CIFAR-10} & \multicolumn{2}{c}{CIFAR-100} & \multicolumn{2}{c}{SVHN} & \multicolumn{2}{c}{Fashion-MNIST} & \multicolumn{2}{c}{Quick, Draw!} \\
 & Regret & Time & Regret & Time & Regret & Time & Regret & Time & Regret & Time\\
\noalign{\smallskip}
\hline
\noalign{\smallskip}
No Stopping & \textbf{0.00} & 295 & \textbf{0.00} & 293 & \textbf{0.00} & 451 & \textbf{0.00} & 301 & \textbf{0.00} & 526 \\
\cite{Domhan2015} & 0.04 & 152 & 0.82 & 138 & 0.16 & 367 & 0.26 & 171 & 0.02 & 285 \\
\cite{Baker2017a} & 0.04 & 38 & \textbf{0.00} & 39 & 0.16 & 74 & \textbf{0.00} & 38 & \textbf{0.00} & 72 \\
\cite{Chandrashekaran2017} & 0.04 & 46 & 0.82 & 14 & \textbf{0.00} & 90 & \textbf{0.00} & 26 & 0.02 & 56 \\
\ourLCP{} & \textbf{0.00} & \textbf{18} & \textbf{0.00} & \textbf{13} & 0.10 & \textbf{26} & \textbf{0.00} & \textbf{17} & 0.02 & \textbf{31} \\
\hline
\end{tabular}
\end{table*}

\subsubsection{Accelerating Random Search}

In this experiment we analyze the impact of extrapolating learning curves when combined with a neural architecture search.
For simplicity, we accelerate a random search being conducted on the search space as described in Section \ref{sub:search-space}.
Every neural architecture is trained for at most 70 epochs and can be terminated early by the learning curve extrapolation methods.
Initially, only one random neural architecture is trained.
Then, sequentially 100 further neural architectures are sampled.
Every learning curve extrapolator iterates over the architectures in the same sequence.
At every epoch, it decides whether to terminate or continue the training process.
We observe that the learning curve extrapolator by \citet{Domhan2015} always chooses to terminate early if only the very first epochs are seen.
Thus, we follow the recommendation of \citet{Domhan2015} and do not terminate before having seen the first 30 epochs.
The results are summarized in Table \ref{tab:random-search}.
First, we observe that all methods accelerate the search with almost no regret, i.e. there is a negligible drop in accuracy  over random search without early stopping (``No Stopping'').
Unsurprisingly, the extrapolator by \citet{Domhan2015} requires the most amount of time, as it is unable to make useful decisions for shorter learning curves.
Furthermore, we can confirm the results by \citet{Baker2017a} and \citet{Chandrashekaran2017} who both claim to be better than \citet{Domhan2015}.
Finally, our method takes the least amount of time, beating all the other methods.
The training time is shortest across all the datasets, nonetheless,  the regret is never significantly higher than any other method.

\section{Discussion of Limitations}

Our neural architecture selection has two limitations we want to discuss.
First, it requires the existence of metaknowledge for the task at hand which can be computational expensive to create.
For our experiments we invested about 1,866 hours GPU hours for its creation.
However, this only needs to be done once and then can be shared with other researchers to reduce the search time significantly.
This is also paying off in the long run for AutoML services where companies offer affordable architecture optimization methods to their customers.
Furthermore, it is not required that this metaknowledge is gathered by a single person.
OpenML.org has proven that collaborative metaknowledge collection works.
We would also like to compare this computational effort to the effort spent by researchers to develop new architectures such as ResNet or DenseNet.
Immense human and computational effort was spent to create these architectures but now they are available for everyone.

Second, our architectures are selected from a smaller candidate pool compared to traditional neural architecture search spaces.
We argue that working on a random subset of the entire search space used by other neural architecture searches is neither an advantage or disadvantage.
The fact that our method outperforms very competitive methods is an evidence that any random subset (if chosen large enough) contains good architectures with high probability.
This claim is supported by most architecture search papers which show and claim that random search is a very competitive baseline (in particular by \citet{Zoph2017a} who introduced this search space; see also Figure \ref{fig:opt-results-short}).
Moreover, it appears that the large search space contains many global optima of equal quality.
Thus, an increase of the search space does not affect the problem's difficulty.
Often it is claimed that the search on a larger space is supposed to be able to find new architectures which can give us more insight.
However, recent architecture search papers use more constrained search spaces which leave no scope for surprises.
Thus, an architecture selection is equally valid as a search on an admittedly larger but very restricted space.
\section{Conclusions}

We present two novel methods for neural architecture optimization (\ourNAS{}) and learning curve extrapolation (\ourLCP{}) based on inductive transfer.
We show that \ourNAS{} achieves state-of-the-art results on CIFAR-10 and CIFAR-100.
With no search time it obtains outstanding results, outperforming all but one competitor method.
Spending 8 hours search time, it is able to outperform the last competitor method ENAS in equal time.
In a more elaborate study on five different datasets, we provide empirical evidence that our method is equal or better than ENAS.
Further advantages over ENAS are that it can profit from early stopping and that it provides strong neural architecture recommendations in no time.

Additionally, we compare our proposed learning curve extrapolation method \ourLCP{} to three competitor methods on five benchmark datasets.
Our extrapolation method is the very first to use inductive transfer in order to improve the forecast.
In an experiment to accelerate random search by means of early stopping, we show significantly faster results with comparable validation accuracy.

In conclusion, we demonstrated that inductive transfer is a useful tool to improve automated machine learning with respect to training time and classification performance.
This is a very interesting finding which might be the start of a new research area.

\bibliography{myrefs}
\bibliographystyle{icml2019}

\end{document}